\documentclass[letterpaper, 10 pt, conference]{ieeeconf}  % Comment this line out if you need a4paper

\IEEEoverridecommandlockouts                              % This command is only needed if 
\usepackage{bm}
\usepackage{graphicx}
\usepackage{amsmath}
\usepackage{amsfonts}
\usepackage{amssymb}
\usepackage{multirow}
\usepackage{wrapfig}
\usepackage{color}

\usepackage{cite}    
\makeatletter
\let\NAT@parse\undefined
\makeatother
\usepackage{hyperref}

\usepackage{algorithm}
\usepackage{algorithmic}
\usepackage{booktabs}

\newtheorem{assumption}{Assumption}
\newtheorem{theorem}{Theorem}

\title{\LARGE \bf
Unified Meta-Representation and Feedback Calibration \\for General Disturbance Estimation
}

\author{Zihan Yang$^{1}$, Jindou Jia$^{2}$, Meng Wang$^{2}$, Yuhang Liu$^{1}$, Kexin Guo$^{1,\dag}$ and Xiang Yu$^{2}$% <-this % stops a space
\thanks{*This work was not supported by any organization}% <-this % stops a space
\thanks{$^{1}$Zihan Yang, Yuhang Liu, and Kexin Guo are with the School of Aeronautic Science and Engineering, Beihang University, Beijing 100191, China}%
\thanks{$^{2}$Jindou Jia, Meng Wang and Xiang Yu are with the School of Automation Science and Electric Engineering, Beihang University, Beijing 100191, China}%
}

\begin{document}

\maketitle
\thispagestyle{empty}
\pagestyle{empty}

%%%%%%%%%%%%%%%%%%%%%%%%%%%%%%%%%%%%%%%%%%%%%%%%%%%%%%%%%%%%%%%%%%%%%%%%%%%%%%%%
\begin{abstract}
Precise control in modern robotic applications is always an open issue due to unknown time-varying disturbances. 
Existing meta-learning-based approaches require a shared representation of environmental structures, 
which lack flexibility for realistic non-structural disturbances. 
Besides, representation error and the distribution shifts can lead to heavy degradation in prediction accuracy. 
This work presents a generalizable disturbance estimation framework that builds on meta-learning and feedback-calibrated online adaptation. 
By extracting features from a finite time window of past observations, 
a unified representation that effectively captures general non-structural disturbances can be learned without predefined structural assumptions. 
The online adaptation process is subsequently calibrated by a state-feedback mechanism to attenuate the learning residual
originating from the representation and generalizability limitations.
Theoretical analysis shows that simultaneous convergence of both the online learning error and the disturbance estimation error can be achieved. 
Through the unified meta-representation, our framework effectively estimates multiple rapidly changing disturbances, 
as demonstrated by quadrotor flight experiments.
See the project page for video, supplementary material and code: \url{https://nonstructural-metalearn.github.io}.
\end{abstract}

%===============================================================================
\section{Introduction}
Modern intricate robotic systems can be confronted with dynamic and complex environments that introduce
unknown disturbances in the nominal case, 
including latent dynamics variation and other external disturbances.
To retain high control precision,
such disturbances must be handled properly.
Several classical methods,
such as Disturbance Observer (DO) \cite{guo_disturbance_rejection_2011},
Incremental Nonlinear Dynamic Inversion (INDI) \cite{nanNonlinearMPCQuadrotor2022} 
and $\mathcal{L}_1$ adaptive control \cite{l1_adapt_ctrl,huangDATTDeepAdaptive2023},
directly estimate the lumped disturbance,
but are limited by the trade-off between disturbance estimation lag and noise amplification.
The utilization of the feedforward model can greatly improve the estimation performance
with a specific type of disturbance.
Multi-model-based estimation of multiple, heterogeneous, and isomeric disturbances is studied in \cite{guo_antidisturbance_2014}
with remarkable results.
Recently, adaptive control with a meta-learning scheme has allowed robots to 
learn from prior experiences and rapidly adapt to new environments 
\cite{clavera2018learning,RSS2021_Adaptive-Control-Oriented,neuralfly}.
Such frameworks enable the rapid estimation of environmental disturbances with shared features,
benefiting from the power of offline learning and online adaptation.

Despite their promising results, several drawbacks are worth mentioning.
Firstly, these methods are designed for disturbances with shared structural representation.
The disturbances are assumed to be functions of the state representation
and an environment configuration,
e.g., the wind disturbances \cite{RSS2021_Adaptive-Control-Oriented, neuralfly}.
Nevertheless, the premise of the existence of shared representation may not hold in a general unstructured environment, 
e.g., sudden external forces or coupled effects with unknown disturbances.
Moreover, some well-conditioned environments
are tricky to construct in real-world data collection for domain-invariant meta-learning.
Secondly, the learning residuals are not explicitly considered in existing frameworks.
The approximation capability of the meta-learning approach is guaranteed by collecting information in various cases 
\cite{fallahGeneralizationModelAgnosticMetaLearning2021, khoeeDomainGeneralizationMetalearning2024}.
However, generalization degrades when the environment lies outside the support of the training task distribution.
While meta-learned representations are trained on disturbances from a specific distribution, 
real-world disturbances may not match this distribution, leading to a performance drop.

\begin{figure*}[t]
  \centering
  \includegraphics[width=0.8\textwidth]{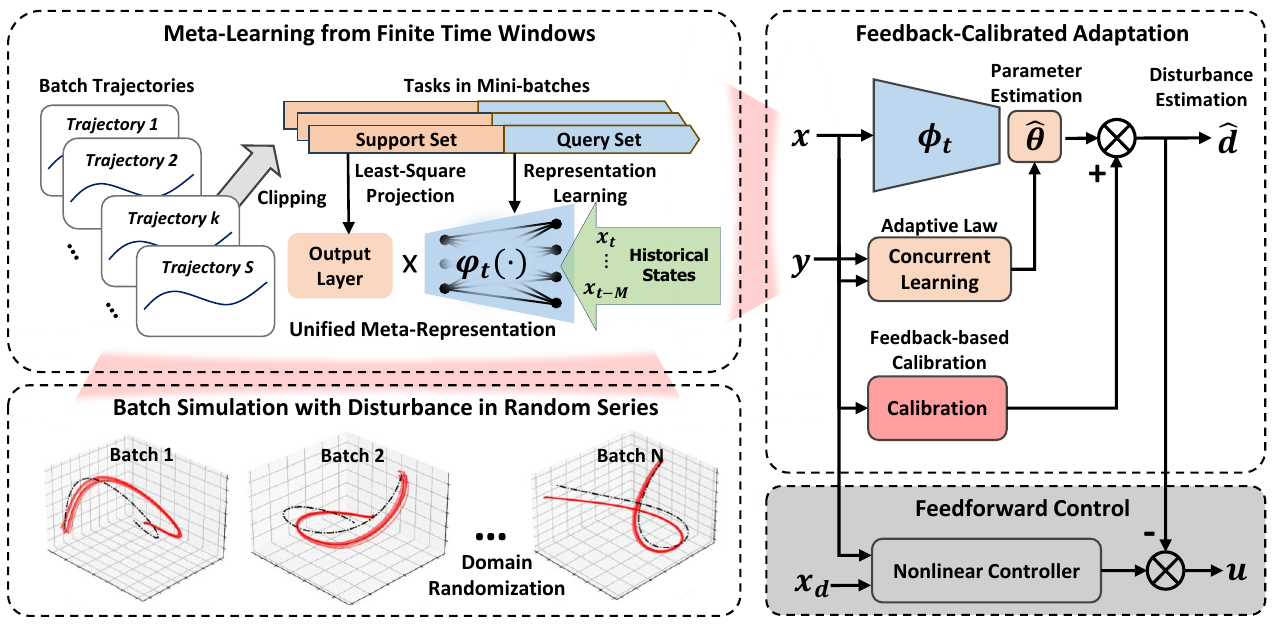}
  \caption{Schematic of the proposed framework. The meta-representation is learned from a finite time window of past observations with 
  domain-randomized disturbances. The online adaptation is calibrated with a feedback mechanism to attenuate the learning residual, which can be
  further integrated with a baseline controller for disturbance rejection.}
  \label{fig:framework}
  \vspace{-1.0em}
\end{figure*}

Motivated by these challenges, we propose a feedback-calibrated meta-adaptive framework to aid generalization
from both the representation and calibration perspectives.
Initially, a unified meta-representation that captures general disturbance effects is constructed 
using a finite time window of past observations, without requiring any predefined structural assumptions.
Domain randomization in simulation generates diverse disturbances using random Fourier series \cite{domain_rand_dnn,dynamics_rand}, 
enabling strong generalization for both simulation and real-world tasks.
Next, a feedback-calibrated mechanism is incorporated to correct prediction errors and attenuate residuals arising 
from representation limitations and generalizability loss, ensuring reliable online adaptation and disturbance estimation.
To demonstrate the effectiveness of the proposed method,
we incorporate the framework with a baseline controller and construct extensive empirical studies
on the trajectory tracking control of a quadrotor under multiple rapid-changing disturbances.
Such disturbances
include aerodynamic drag, unknown external forces, unknown fixed and suspended payloads,
as well as external wind disturbances.
The results indicate that using a unified meta-representation with online feedback calibration can generalize to such non-structural disturbances,
also with significant improvements over the state-of-the-art approaches.
The main contributions of this work can be summarized as follows:
\begin{itemize}
  \item A unified meta-representation for modelling general non-structural disturbances
  without predefined structural assumptions.
  \item A feedback-calibrated mechanism for the attenuation of the meta-adaptation residual, 
  thereby improving the model accuracy and generalizability.
  \item Theoretical analysis for the simultaneous convergence of the online learning error and the disturbance estimation error.
\end{itemize}

%===============================================================================
\section{Related Works}
% meta-learning的方式与区别
\subsection{Offline Meta-Learning}
Meta-learning aims at learning a representation from previous experiences that can 
adapt to new tasks quickly \cite{pmlr-v70-finn17a, metalearn_nn}.
In the context of model-based control, meta-learning has been applied to
adaptive control \cite{neuralfly,RSS2021_Adaptive-Control-Oriented,icra2024_HMAC}
and model-based reinforcement learning (MBRL) \cite{ModelBasedMetaReinforcementLearning2021c,clavera2018learning} 
with promising results.
Meta-learners can be designed in various ways, including optimizing hyperparameters of the base-learner
and/or learning a representation (usually a neural network) for the adaptation task of the base-learner.
% Here we review the latter case for the sake of relativity.
A domain-adversarially invariant approach is proposed to 
learn a shared spatial representation of different wind disturbances with constant wind speeds \cite{neuralfly}.
A control-oriented method is presented for end-to-end learning of the adaptation policy \cite{RSS2021_Adaptive-Control-Oriented}.
Model predictive control (MPC) with adaptation is constructed 
for fast adaptation to new conditions \cite{metalearn_RecedingHorizon}.
Learning representations with structures 
from bilinear models to deep neural networks (DNNs) has been studied \cite{NEURIPS2021_52fc2aee}.
An intriguing approach learns to adapt the controller
gains concerning various environments, which is augmented with a Kalman filter for parameter learning
\cite{sanghviOCCAMOnlineContinuous2024}.
% The above methods fall into the category of 
% linear coefficient adaptation and nonlinear representation learning.
% Test-Time task inference with MBRL is applied in \cite{ModelBasedMetaReinforcementLearning2021c} 
% for suspended payload adaptation.
% Self-supervised full-layer DNN adaptation is further developed based on MAML
% framework with stability guarantees \cite{he2024selfsupervisedmetalearningalllayerdnnbased}.
Despite the effectiveness of structural disturbances, 
these methods require a shared spatial representation among varied disturbances or tasks, 
which is unsuitable for non-structural disturbances,
e.g., the joint effects of external gust disturbance and unknown payload.
A framework that refines the tasks into previous trajectory segments and 
learns both the base-learner hyperparameter and the representation has been proposed \cite{clavera2018learning},
enabling MBRL on different failure modes and loading conditions 
of legged robots and manipulators.
A hierarchical framework \cite{icra2024_HMAC} that enables the capture of unmanageable environmental disturbances
can alleviate such an issue.
Our method also focuses on representation learning 
and handles the latent disturbances in a smooth and streaming way, 
but we show that it is possible to yield a unified representation, based on finite-time observations and domain randomization,
that significantly improves the model performance. 
To alleviate the difficulties of data collection for meta-learning in the previous methods, 
we proposed a simulation-based domain randomization
to collect various kinds of non-structural disturbances for representation learning.

\vspace{-0.5em}
\subsection{Online Adaptation}
In the online adaptation phase of meta-learning, the model parameter is updated with upcoming data.
In the context of model-based control, online adaptation is usually achieved by adaptive laws
\cite{slotineCompositeAdaptiveControl1989a, annaswamyHistoricalPerspectiveAdaptive2021}
in adaptive controllers and stochastic gradient descent \cite{clavera2018learning, pmlr-v70-finn17a, active_learn_mpc} 
in MBRL or MPC approaches.
The composite adaptive law consists of multiple sources of feedback to boost the convergence
\cite{RSS2021_Adaptive-Control-Oriented, neuralfly, icra2024_HMAC}.
In real-world applications, the utilization of Kalman filters \cite{neuralfly,sanghviOCCAMOnlineContinuous2024}
enables parameter learning from noisy measurements.
However, none of the existing adaptation methods explicitly considered the learning residual that
can originate from representation error and distribution shifts.
We propose a feedback-calibrated mechanism that attenuates the learning residual
that favors the final estimation performance.

% \textbf{Other Online Learning-based Disturbance Estimators.}
% Besides the meta-learning-based approaches,
% some online learning-based methods are worth mentioning.
% Online learning-based disturbance observers 
% are proposed in \cite{jiaEVOLVEROnlineLearning2024,jiaDisturbanceObserverEstimating2024}
% for fast and accurate disturbance estimation.
% A Gaussian process-based L1-adaptive control is proposed in \cite{gahlawat2020_L1gp}. 
\begin{figure*}[t]
  \centering
  \includegraphics[width=0.9\textwidth]{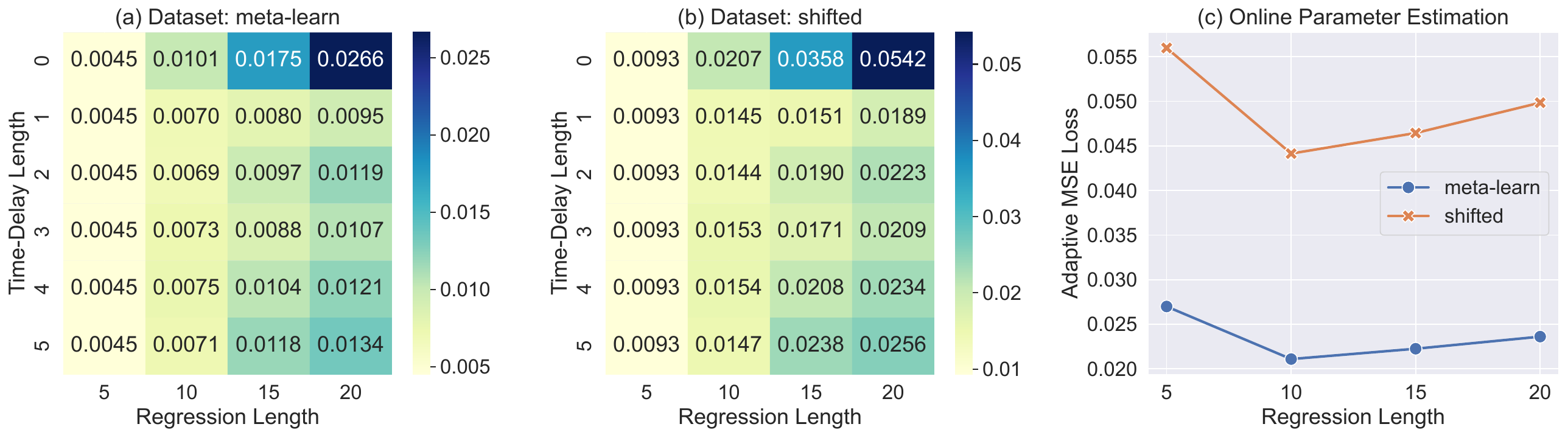}
  \caption{The result of ablation study, including model performance (prediction loss in mean squared error) on \textit{meta-learn} dataset (a),
  \textit{shifted} dataset (b) and the effect of online parameter estimation (c).}
  \label{fig:ablation}
  \vspace{-1.0em}
\end{figure*}

%===============================================================================
\section{Problem Formulation}
\label{sec:problem}
We consider a control affine system, a general case for rigid-body robots:
\begin{equation}
  \dot{\bm{x}} = \bm{f}(\bm{x}) + \bm{g}(\bm{x})\bm{u} + \bm{d}(\bm{x}, \bm{h}(t))
\end{equation}
where $\bm{x} \in \mathbb{R}^n$ and $\dot{\bm{x}} \in \mathbb{R}^n$ are the state and its derivative, respectively.
$\bm{u} \in \mathbb{R}^m$ is the control input and
$\bm{f}(\bm{x})$ and $\bm{g}(\bm{x})$ are continuously differentiable mappings.
The disturbance $\bm{d}(\cdot) \in \mathbb{R}^n$ originates from state-related internal effects and
external time-varying environmental impact $\bm{h}(t) \in \mathbb{R}^h$.
Our key objective is to represent general $\bm{d}(\bm{x}, \bm{h}(t))$ based on a unified meta-representation
that covers the influence of $\bm{h}(t)$ and can adapt to new disturbances.
As shown in \cite{RSS2021_Adaptive-Control-Oriented,neuralfly,icra2024_HMAC},
we formulate the disturbance as:
\begin{equation}
  \label{eq:disturbance}
  \bm{d}(\bm{x}, \bm{h}(t)) = \bm{\Xi} \bm{\varphi}(\bm{x}, \bm{h}(t)) + \bm{\epsilon}
\end{equation}
where $\bm{\Xi} \in \mathbb{R}^{n \times k}$ a general online-learned model parameter,
$\bm{\varphi}(\cdot): \mathbb{R}^n \times \mathbb{R}^h \rightarrow \mathbb{R}^k$ 
is the shared representation among different disturbance structures.
$\bm{\epsilon} \in \mathbb{R}^{n}$ refers to the learning residual, indicating that the formulation is imperfect under
the representation error and model generalizability loss.
Previous approaches assume that $\bm{h}(t)$ is a constant or slow-varying environment configuration represented by $\bm{\Xi}$,
such as wind speed or payload mass, then $\bm{d}(\bm{x}, \bm{h}(t))$ is approximated with the formulation 
$\bm{\Xi} \bm{\varphi}(\bm{x})$. This limits the representation in a structured pattern characterized by $\bm{\varphi}(\bm{x})$.

Distinguishing the previous methods \cite{RSS2021_Adaptive-Control-Oriented,neuralfly,icra2024_HMAC}, 
which extracts a global, state-dependent structure 
from the entire disturbance space, our method learns a unified shared 
representation from a finite time window of state observations.
% $\bm{h}(t)$ is included in the representation $\bm{\varphi}(\cdot)$ to enhance the approximation capability, 
% thereby preserving temporal variability and enabling flexible 
% adaptation to general non-structural disturbances.
We propose a unified representation for unstructured disturbances that encodes a sequence of previous states to 
implicitly address the influence of $\bm{h}(t)$ in $\bm{\varphi}(\cdot)$, thereby preserving temporal variability and enabling flexible 
adaptation to general non-structural disturbances.
% where the injection of disturbances
% is indicated by the sequential state variation in the time window.
Additionally, a feedback-calibrated mechanism is introduced to handle the learning residual
to enhance the model accuracy and generalizability.

%===============================================================================
\begin{figure}[t!]
  \centering
  \includegraphics[width=0.9\linewidth]{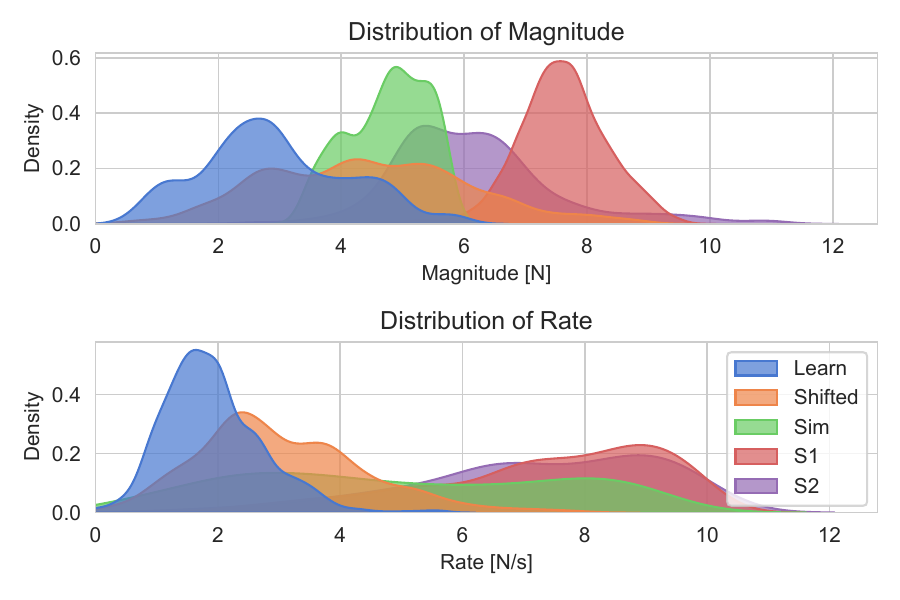}
  \vspace{-0.5em}
  \caption{Distribution differences of disturbances in the dataset of \textit{meta-learn}
  (Learn), \textit{shifted} (Shifted), simulations (Sim) and real-world scenarios (S1 and S2).}
  \vspace{-1.0em}
  \label{fig:distribution}
\end{figure}

\section{Methods}
\subsection{Meta-learning for Non-Structural Disturbances}
For simplicity and clarity, we rewrite the formulation $\bm{\Xi} \bm{\varphi}(\cdot)$ as
$\bm{\phi}(\cdot)\bm{\xi}$, where $\bm{\xi} = [\bm{\Xi}_1^\top,\cdots,\bm{\Xi}_k^\top]^\top \in \mathbb{R}^{nk}$ 
is the concatenated columns of $\bm{\Xi}$,
and $\bm{\phi}(\cdot) = \text{diag} \{\bm{\varphi}(\cdot)^\top, \cdots, \bm{\varphi}(\cdot)^\top\} \in \mathbb{R}^{n \times nk}$ 
is the diagonal concatenation of $\bm{\varphi}(\cdot)^\top$.

\subsubsection{Unified Meta-Representation}
Given a disturbance model in the form of \eqref{eq:disturbance}, we aim to learn a meta-representation for
$\bm{\phi}(\bm{x}, \bm{h}(t))$ without the knowledge of $\bm{h}(t)$.
Inspired by time-delay embedding \cite{10.1007/BFb0091924}, we propose a meta-learned representation
that encodes a historical state segment into a latent space for disturbance prediction.
Existing approaches have shown that the time-delay embedding can be utilized for disturbance and dynamics modeling
\cite{online_dynamics,piqueControllingSoftRobotic2022}.
The disturbance model is reformulated as:
\begin{equation}
  \bm{d}(\bm{x}, \bm{h}(t)) = \bm{\phi}(\bm{x}, \bm{h}(t))\bm{\xi}  + \bm{\epsilon} = \bm{\phi}_t({\bm{z}})\bm{\theta}  + \bm{\gamma}
\end{equation}
where $\bm{z}_k = [\bm{x}_k, \bm{x}_{k-1}, \cdots, \bm{x}_{k-M}]$ and $M$ is the embedding dimension.
$\bm{\theta} \in \mathbb{R}^{nk}$, $\bm{\phi}_t(\cdot): \mathbb{R}^{nM} \rightarrow \mathbb{R}^{n\times nk}$ 
and $\bm{\gamma} \in \mathbb{R}^n$ 
are the new model parameter, the meta-learned representation 
and learning residual under time-delay embedding, respectively.
% Such model can be augmented with the nominal dynamics to form the meta-learned dynamics:
% $\dot{\bm{x}} = \bm{f}(\bm{x}) + \bm{g}(\bm{x})\bm{u} + \bm{\phi}_t(\bm{z})\bm{\theta}$.
% Next, we show how to learn $\bm{\phi}_t(\bm{z})$ with meta-learning.

\subsubsection{Learning from Finite Time Window}
Generally, the meta-learning problem is achieved by bi-level optimization.
The inner base-learning problem learns the model parameter for adaptation
based on the task-specified support set $\mathcal{D}_{support}$ and a given representation.
The outer meta-learning problem learns an optimal representation
for the adaptation to new tasks (query set) $\mathcal{D}_{query}$.
The meta-learning algorithm
is designed for future prediction based on past data-based adaptation.
In particular, a task $\mathcal{D}$ is defined as multiple trajectory segments
$\mathcal{D}=\{\mathcal{D}^{1}, \cdots, \mathcal{D}^{N_m}\}$,
where $\mathcal{D}^{i}=\{\bm{x}_{1:N+M+H}^{i}, \bar{\bm{d}}_{1:N+M+H}^{i}\}$ is the state-disturbance sequence.
$\bar{\bm{d}}$ is the real disturbance that can be obtained from simulation or offline-filtered real-world data.
$\mathcal{D}^{i}_{support}$ and $\mathcal{D}^{i}_{query}$ are divided from $\mathcal{D}^{i}$
and refer to the past $N$ and future $H$ state-disturbance sequences, respectively.
The base-learner is a regularized least-square method that learns the model parameter $\bm{\theta}^*$
with a given $\bm{\phi}_t$ from past $N$ state-disturbance sequence.
We note that $\bm{\phi_t}(\cdot)$ is parameterized by $\bm{\eta}$.
For the meta-learner, the objective is to learn an optimal $\bm{\eta}$ with the
base-learned $\bm{\theta}^*$ for the prediction of future $H$ disturbances:
% \begin{equation}
%   \label{eq:meta-learning}
%   \begin{aligned}
%     \bm{\eta}^* \in \;
%     &\arg\min_{\bm{\eta}} \frac{1}{H}\sum_{k=1}^H \frac{1}{2}\|{\bm{\phi}_t(\bm{z}_k) \bm{\theta}^* - \bar{\bm{d}}_k}\|^2_1 
%     + \lambda_1\|{\bm{\phi}_t}\|_1,\; \bar{\bm{d}}_k \in \mathit{D_{query}^{i}}\\
%     &\quad\quad s.t.\;
%     \bm{\theta}^* =
%     \arg\min_{\bm{\phi}_t} \frac{1}{N}\sum_{k=1}^N \frac{1}{2}\|{\bm{\phi}_t(\bm{z}_k)\bm{\theta} - \bar{\bm{d}}_k}\|^2_2 
%     + \lambda_2\|{\bm{\theta}}\|^2_2 ,\; \bar{\bm{d}}_k \in \mathit{D_{support}^{i}}\\
%     \end{aligned}
% \end{equation}
\begin{equation}
  \label{eq:meta-learning}
  \begin{aligned}
    \min_{\bm{\eta}}&\;\frac{1}{H}\sum_{k=1}^H \frac{1}{2}\|{\bm{\phi}_t(\bm{z}_k) \bm{\theta}^* - \bar{\bm{d}}_{kU}}\|^2_1 
    + \lambda_1\|{\bm{\phi}_t}\|_1,\\
    \text{s.t.}&\;
    \bm{\theta}^* =
    \arg\min_{\bm{\theta}} \frac{1}{N}\sum_{k=1}^N \frac{1}{2}\|{\bm{\phi}_t(\bm{z}_k)\bm{\theta} - \bar{\bm{d}}_{kL}}\|^2_2 
    + \lambda_2\|{\bm{\theta}}\|^2_2.
  \end{aligned}
\end{equation}
where $\bar{\bm{d}}_{kU} \in \mathit{D_{query}^{i}}$ and $\bar{\bm{d}}_{kL} \in \mathit{D_{support}^{i}}$.
$\lambda_1$, $\lambda_2$ are the L1 and L2 regularization parameters that encourage simpler representation.
The inner problem is with the closed-form solution:
\begin{equation}
  \label{eq:inner_problem}
  \bm{\theta}^* = (\bm{\Phi}^\top \bm{\Phi} + \lambda_2 \bm{I})^{-1} \bm{\Phi}^\top \bar{\bm{\Delta}}
\end{equation}
where $\bm{\Phi} = [\bm{\phi}_t(\bm{z}_1), \cdots, \bm{\phi}_t(\bm{z}_N)]^\top$ and
$\bar{\bm{\Delta}} = [\bar{\bm{d}_1}, \cdots, \bar{\bm{d}_N}]^\top$ are the concatenated vectors for regression.
The original bi-level optimization problem \eqref{eq:meta-learning} is therefore a single layer one.
% \begin{equation}
%   \label{eq:meta-learning_single}
%   \bm{\eta}^* \in \;
%   \arg\min_{\bm{\eta}} \frac{1}{H}\sum_{k=1}^H 
%   \frac{1}{2}\|{\bm{\phi}_t(\bm{z}_k)
%   (\bm{\Phi}^\top \bm{\Phi} + \lambda_2 \bm{I})^{-1} \bm{\Phi}^\top \bar{\bm{\Delta}} - \bar{\bm{d}}_k}\|^2_2 
%   + \lambda_1\|{\bm{\phi}_t}\|_1
% \end{equation}

\begin{algorithm}[H] 
\caption{Meta-Learning from Segments}
\label{learning algorithm}
  \begin{algorithmic}[1]
    \REQUIRE{Base-learner regression size $N$, time-delay-embedding size $M$,
      mini-batch size $N$; dataset $\mathcal{D}$, objective function $J$ in \eqref{eq:meta-learning}.}
    \ENSURE{Representation $\bm{\phi}_t$}. \\
    \COMMENT\ {Representation parameters $\bm{\eta}$; 
    slice $\mathcal{D}$ into $N_m$ segments $\mathcal{D}=\{\mathcal{D}^{1}, \cdots, \mathcal{D}^{N_m}\}$ with length $H$ each,
    $\mathcal{D}^{i}=\{\bm{x}_k^{i}, \bar{\bm{d}}_k^{i}\},\;i=1,\cdots,N_m$.
    }
    \REPEAT 
    \FOR{$\{\bm{x}_{1:H}^{i}, \bar{\bm{d}}_{1:H}^{i}\}$ in $\{\mathcal{D}_{i = 1, \cdots, N_m}\}$}
    \STATE Compute Gradient $\nabla_{\bm{\eta}}J$ based on \eqref{eq:meta-learning};
    \STATE Compute step $\Delta \bm{\eta}$ using \textit{Adam} or other methods;
    \STATE Update $\bm{\eta}$ with $\bm{\eta} \leftarrow \bm{\eta} + \Delta \bm{\eta}$;
    \ENDFOR
    \UNTIL{\textbf{convergence}}
  \end{algorithmic}
\end{algorithm}

% The problem is non-convex but allows $\bm{\phi}_t$ to be trained using conventional stochastic gradient descent methods.
The mini-batching technique is applied for rolling out all trajectory segments with gradient descent on the parameter of $\bm{\phi}_t$
based on the bi-level optimization \eqref{eq:meta-learning}. 
Since it is defined for a single trajectory segment $\mathcal{D}^{i}$,
mini-batch learning can be applied to the complete dataset $\mathcal{D}$.
The complete learning phase is shown in Algorithm \ref{learning algorithm}.
The real-time computation of the least-square method with large parameter dimensions can be heavy for limited-sourced robots.
Therefore, the $\bm{\theta}^*$ is estimated with adaptive laws in the online adaptation phase.
More details on the meta-learning can be found in the
\href{https://nonstructural-metalearn.github.io/static/pdfs/apdx.pdf}{\textcolor{blue}{Supplementary Material I-C}}.

\subsubsection{Ablation Study on Model Parameters}
% To validate the effectiveness of the spatial-temporal representation,
% we conduct an ablation study on the model parameters as well as how the online adaptation affects the model performance.
Two datasets are constructed for the ablation study.
Besides \textit{meta-learn} set for training, \textit{shifted} set is loaded
with disturbances containing larger magnitudes and faster rates compared with that of \textit{meta-learn}.
As shown in Figure.\ref{fig:ablation}, the time-delay embedding with $M=3$ outperforms the one with $M=1$ in both datasets
and further improves as the regression length $N$ increases.
While smaller $N$ leads to better model accuracy, the online performance is degraded
if online parameter estimation \eqref{eq:online_adaptation} is applied instead
of directly solving the least-square problem \eqref{eq:inner_problem}. A trade-off 
between the convergent speed and the model performance can be obtained with $N=10$.
More details on the meta-learned models can be found in the
\href{https://nonstructural-metalearn.github.io/static/pdfs/apdx.pdf}{\textcolor{blue}{Supplementary Material I-A}}

\subsubsection{Domain Randomization for Meta-learning}
To generalize to arbitrary non-structural disturbances, $\mathcal{D}$ is constructed by simulations.
Domain randomization \cite{domain_rand_dnn,dynamics_rand} 
is applied in a batch simulator to construct non-structural disturbances
in random series with different magnitudes and rates under various closed-loop conditions.
Despite the abundance of simulated cases, the model performance
can be damaged by the representation error and distribution shifts.
In the next section, we introduce a feedback-calibrated mechanism that tackles such issues
while doing online parameter estimation. 
See the \href{https://nonstructural-metalearn.github.io/static/pdfs/apdx.pdf}{\textcolor{blue}{Supplementary Material I-B}} for more information.

\begin{figure*}[t!]
  \centering
  \includegraphics[width=0.8\linewidth]{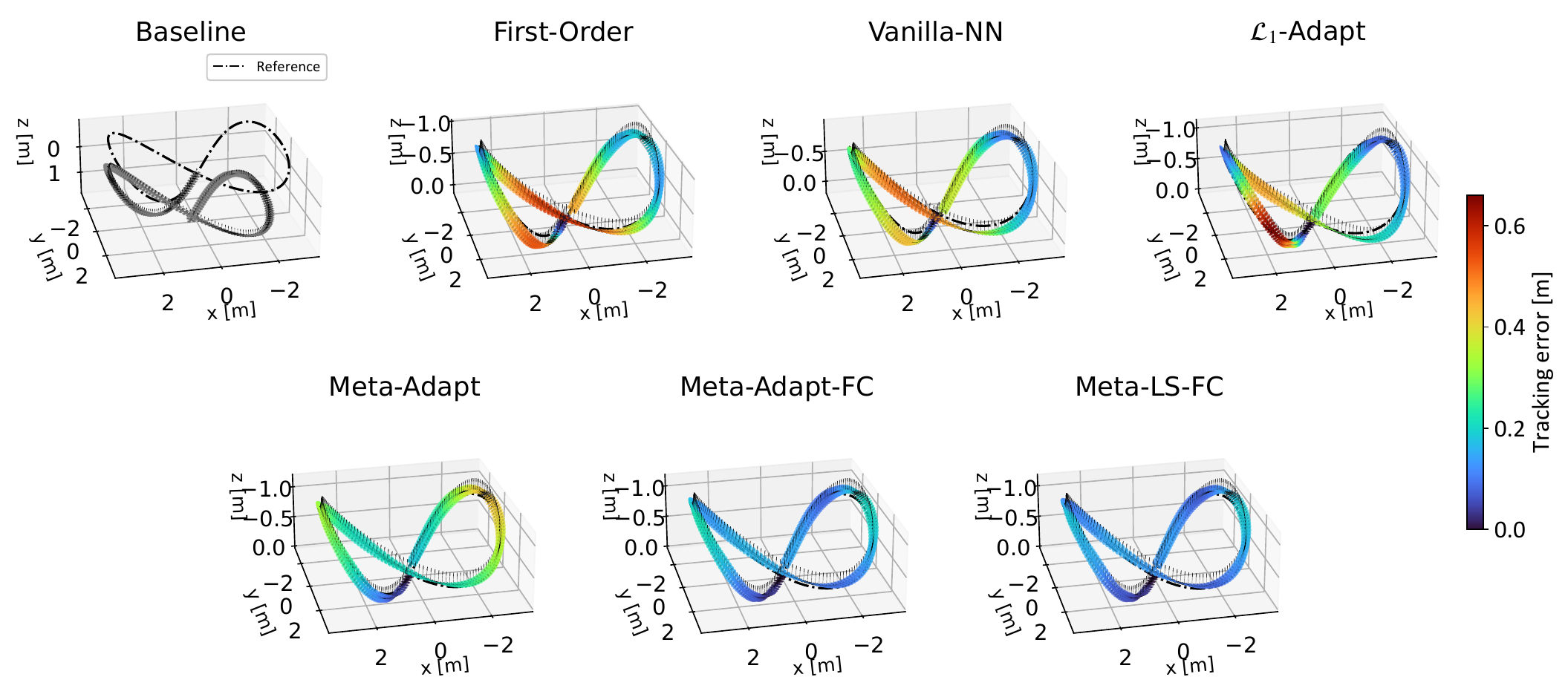}
  \caption{Trajectory tracking results of the simulated cases.}
  \label{fig:sim_ctrl}
\end{figure*}

\subsection{Feedback-Calibrated Online Adaptation}
In attempt to tamp the learning residual $\bm{\gamma}$, 
our framework is furthermore augmented with the feedback-calibration phase by online learning the model parameter $\bm{\theta}$.
To construct the online adaptation algorithm, we start with a feedback-based correction that can be seen in the design
of luenberger observers \cite{Luenberger_obs} and disturbance observers 
\cite{guo_disturbance_rejection_2011}.
\begin{equation}
  \label{eq:dot_feedback}
  \dot{\hat{\bm{d}}} = \dot{\bm{d}}_{model} + \bm{L}(\bm{d} - \hat{\bm{d}})
\end{equation}
where $\bm{L}$ is the positive-definite feedback gain, $\hat{\bm{d}}$ is the estimated disturbance,
$\dot{\bm{d}}_{model}$ refers to the derivative of the disturbance model $\bm{\phi}_t(\bm{z})\bm{\theta}$
and $\bm{d}$ is the real disturbance. 
With $\bm{d} = \dot{\bm{x}} - \bm{f}(\bm{x}) - \bm{g}(\bm{x})\bm{u}$, the feedback-calibrated mechanism
is designed as:
% Subsequently, an auxiliary variable $\bm{\xi}$ is introduced to avoid the usage of state derivatives:
% $\bm{\xi} = \hat{\bm{d}} - \bm{d}_{model} - \bm{Lx}$, which yields our feedback-calibrated mechanism:
\begin{equation}
  \begin{aligned}
    \dot{\bm{\xi}} &= -\bm{L}(\bm{\xi} + \bm{d}_{model} + \bm{Lx} + \bm{f}(\bm{x}) + \bm{g}(\bm{x})\bm{u})\\
    \hat{\bm{d}} &= \bm{d}_{model} + \bm{Lx} + \bm{\xi}
  \end{aligned}
  \label{eq:feedback_calibration}
\end{equation}
where $\bm{\xi} \in \mathbb{R}^n$ is an auxiliary variable to avoid the usage of state derivative.
Such calibration brings us the following benefit with the assumption on the learning residual.
\begin{assumption}
  \label{assumption_1}
  In the disturbance model $\bm{d} = \bm{\phi}_t(\bm{z})\bm{\theta}^* + \bm{\gamma}$,
  the learning residual $\bm{\gamma}$ is bounded smooth, i.e., $\| \bm{\gamma} \| \leq \bar{\gamma}$
  and $\| \dot{\bm{\gamma}} \| \leq \bar{d_\gamma}$,
  where $\bm{\theta}^*$ is the optimal model parameter, $\bar{\bm{\gamma}} > 0$ and $\bar{\bm{d_\gamma}} > 0$ are positive constants.
\end{assumption}
\begin{theorem}
  \label{theorem_1}
  Under Assumption \ref{assumption_1}, with $\bm{\theta} = \bm{\theta}^*$,
  the disturbance estimation error $\tilde{\bm{d}} = \hat{\bm{d}} - \bm{d}$ exponentially converges to
  a bounded set regularized by $\bm{L}$ and $\bar{d_\gamma}$. 
\end{theorem}
$\textit{Proof.}$ refers to the \href{https://nonstructural-metalearn.github.io/static/pdfs/apdx.pdf}{\textcolor{blue}{Supplementary Material II-A}}.
The learning residual $\bm{\gamma}$ is therefore tamped by the feedback-based correction mechanism,
reducing to $\bar{d_\gamma}$ instead of $\bar{\gamma}$.
The knowledge of $\dot{\bm{x}}$ with integration in \eqref{eq:feedback_calibration} 
comes with the accessibility of real-time $\bm{d}$ feedback with $\bm{x}$.

As illustrated in adaptive control approaches \cite{RSS2021_Adaptive-Control-Oriented, neuralfly},
the estimation model parameter $\bm{\theta}$ is achieved using concurrent learning adaptive laws \cite{concur_learn, concur_learn_phd}:
\begin{equation}
  \label{eq:online_adaptation}
  \dot{\hat{\bm{\theta}}} = -\bm{P} \sum_{i=1}^{N_c} \bm{\phi}_t(\bm{z}_i)^\top (\bm{d}_i - \bm{\phi}_t(\bm{z})\hat{\bm{\theta}})
  + \bm{\Gamma}\bm{\phi}_t(\bm{z})^\top (\bm{x}_d - \bm{x})
\end{equation}
where $\bm{P}$, $\bm{\Gamma}$ are the positive-definite gains, $N_c$ is the number of concurrent learning samples,
$\bm{d}_i$ is the disturbance measurement for $\bm{\theta}$-estimation.
In real-world applications, Kalman filter-based estimation \cite{neuralfly} 
or other direct filtering methods can be used to
address the noise of $\bm{d}_i$.
An assumption can be made for the online adaptation.
\begin{assumption}
  \label{assumption_2}
  The optimal model parameter $\bm{\theta}^*$ is slow time-varying, i.e., $\| \dot{\bm{\theta}^*} \| \leq \bar{d_\theta}$.
  The representation is bounded in its magnitude and derivative, i.e., $\| \bm{\phi}_t(\bm{z}) \| \leq \bar{\phi}$,
  $\| \dot{\bm{\phi}_t}(\bm{z}) \| \leq \bar{d_\phi}$, 
  $\bar{d_\theta} > 0$ and $ \bar{d_\theta} > 0$ are positive constants.
\end{assumption}
With saturation functions, $\| \bm{\phi}_t(\bm{z}) \| \leq \bar{\phi}$ can be easily satisfied.
Theorem \ref{theorem_2} holds for online parameter learning and disturbance estimation without feedforward control 
i.e. $\bm{\Gamma} = \bm{0}$.
\begin{theorem}
  \label{theorem_2}
  Under Assumption \ref{assumption_1} and Assumption \ref{assumption_2}, 
  both the disturbance estimation error $\tilde{\bm{d}} = \bm{\hat{d} - d}$
  and the parameter estimation error $\tilde{\bm{\theta}} = \bm{\hat{\theta} - \theta}^*$
  exponentially converges to
  bounded sets.
\end{theorem}
$\textit{Proof.}$ refers to the \href{https://nonstructural-metalearn.github.io/static/pdfs/apdx.pdf}{\textcolor{blue}{Supplementary Material II-B}}.
With a general feedback control law with $\bm{g}(\bm{x})$ in full-rank:
$\bm{u} = \bm{g}(\bm{x})^{-1}(-\bm{f}(\bm{x}) + \dot{\bm{x}_d} + \bm{K}(\bm{x} - \bm{x}_d) - \hat{\bm{d}})$
and positive-definite gains of $\bm{\Gamma}$ and $\bm{K}$,
the closed-loop system also ends up with asymptotic stability.
The full-rank assumption on $\bm{g}(\bm{x})$ indicates that the system is fully-actuated,
but for underactuated systems, the feedback control can be achieved using cascade controllers.
Here we skip the proof since it is similar to the proof of Theorem \ref{theorem_2}.

%===============================================================================
\section{Empirical Study}
\subsection{Simulated Experiments}
In this part, the proposed framework is validated in a simulated quadrotor under mass uncertainty and aerodynamic drag.
From Figure.\ref{fig:distribution}, the disturbances covered for learning is insufficient
to represent the testing disturbances, yet the model is expected to generalize well due to
the unified representation and feedback-calibrated mechanism.
The quadrotor dynamics, controller configuration, disturbances, model parameters,
and online adaptation settings can be found in the
\href{https://nonstructural-metalearn.github.io/static/pdfs/apdx.pdf}{\textcolor{blue}{Supplementary Material III}}.

The abbreviations and explanations of the compared methods are as follows.
First-order disturbance observer (\textbf{First-Order}) \cite{guo_disturbance_rejection_2011}
can be seen as a special case of the feedback-calibration mechanism, where $\dot{\bm{d}}_{model} = \bm{0}$.
Neural-network augmented disturbance observer (\textbf{Vanilla-NN}) is
embedded with learned the aerodynamic effects $\bm{R}\bm{D}\bm{R}^\top \bm{v}$ that favors the disturbance estimation.
$\mathcal{L}_1$-Adaptive Control (\textbf{$\mathcal{L}_1$-Adapt}) \cite{huangDATTDeepAdaptive2023} estimates the disturbance
via velocity feedback and a low-pass filter.
\textbf{Meta-Adapt} refers to the proposed meta-learned model with a classic composite adaptive law as in \cite{neuralfly}.
\textbf{Meta-Adapt-FC} refers to the proposed meta-learned model with the feedback-calibrated online adaptation.
\textbf{Meta-LS-FC} has the model parameter that is directly optimized by the base-learner \eqref{eq:inner_problem}
while enabling feedback calibration.
\begin{table}[htbp]
  \centering
  \caption{RMSE of Disturbance Estimation ($m/s^2$) and Trajectory Tracking Control ($m$).}
  \renewcommand\arraystretch{1.2}
  \label{tab:sim_result}
  \scalebox{1.0}{
    \begin{tabular}{l c c}
      \toprule[0.3mm]
      \textbf{Method} & \textbf{Estimation Loss} & \textbf{Control Loss}\\
      \hline
      First-Order & 0.738 & 0.209 \\
      Vanilla-NN & 0.493 & 0.199 \\
      $\mathcal{L}_1$-Adapt & 0.799 & 0.167 \\
      Meta-Adapt & 0.245 & 0.158 \\
      Meta-Adapt-FC & \textbf{0.159} & \textbf{0.083} \\
      Meta-LS-FC & \textbf{0.151} & \textbf{0.074} \\
      \bottomrule[0.3mm]
    \end{tabular}
  }
\end{table}

\begin{figure*}[t!]
  \centering
  \includegraphics[width=1.0\textwidth]{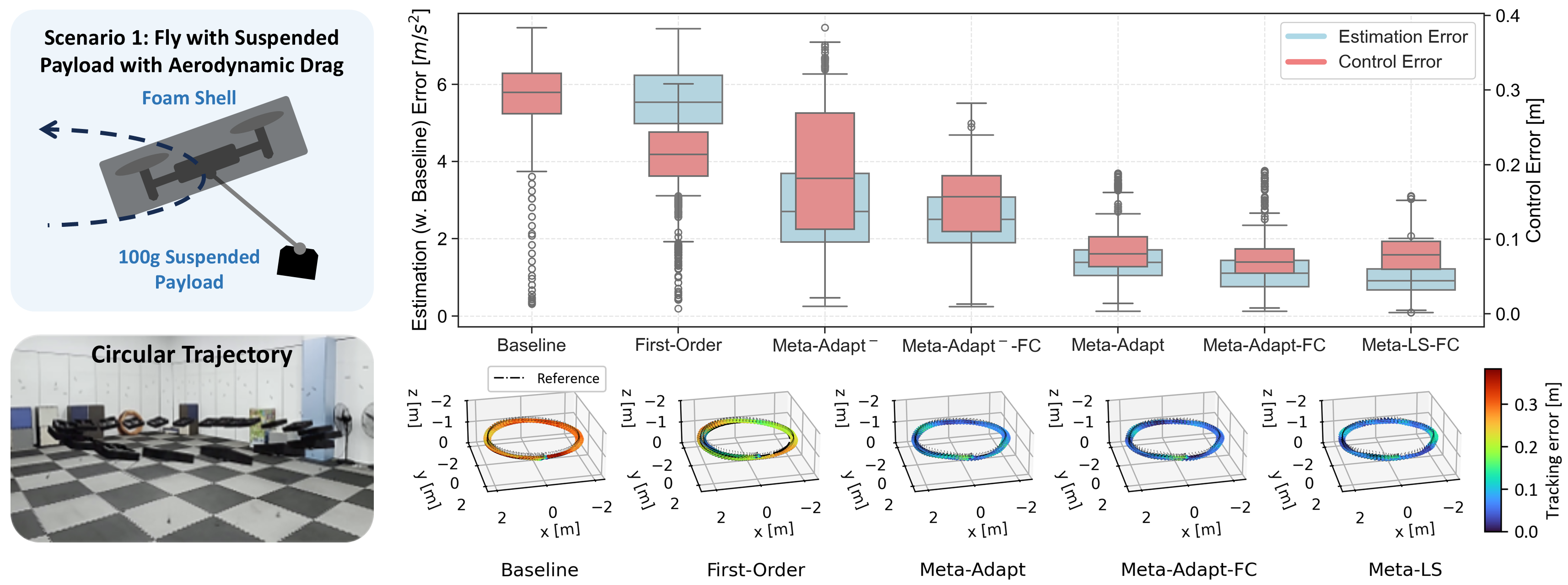}
  \vspace{-1.0em}
  \caption{Scenario.1, the quadrotor maneuvers in circular trajectory with suspended payload and aerodynamic drag. 
  Boxplots of both estimation error and tracking error are provided. 3D trajectories are colored by the tracking error.}
  \vspace{-1.0em}
  \label{fig:exp1_general}
\end{figure*}

For control tasks, the estimators are incorporated into the translational loop of a DFBC 
\cite{mellingerMinimumSnapTrajectory2011,morrellDifferentialFlatnessTransformations2018a} 
baseline controller and servers as a feedforward compensation on the desired acceleration.
For estimation tasks, the estimator works with the baseline without feedforward compensation.
The results are evaluated by root-mean-square error (RMSE) in Table.\ref{tab:sim_result}.
The baseline controller fails under the disturbances with the tracking RMSE of $0.705$.
\textbf{Vanilla-NN} outperforms the \textbf{First-Order} and \textbf{$\mathcal{L}_1$-Adapt} in estimation
since the aerodynamic drag model is captured accurately.
As shown in Figure.\ref{fig:sim_ctrl}, our adaptive approaches significantly boost the estimation and control performance.
With the least-square method \textbf{Meta-LS-FC}, the online adaptation is achieved without the convergent process, which
ends with the highest estimation and tracking control performance.
The feedback-calibration mechanism provides model correction,
leading to an enhancement of $47.5\%$ in the control task and $35.1\%$ in estimation.
Disturbance estimation plots can be found in the
\href{https://nonstructural-metalearn.github.io/static/pdfs/apdx.pdf}{\textcolor{blue}{Supplementary Material III-D}}.

\begin{figure}[t]
  \centering
  \includegraphics[width=1.0\linewidth]{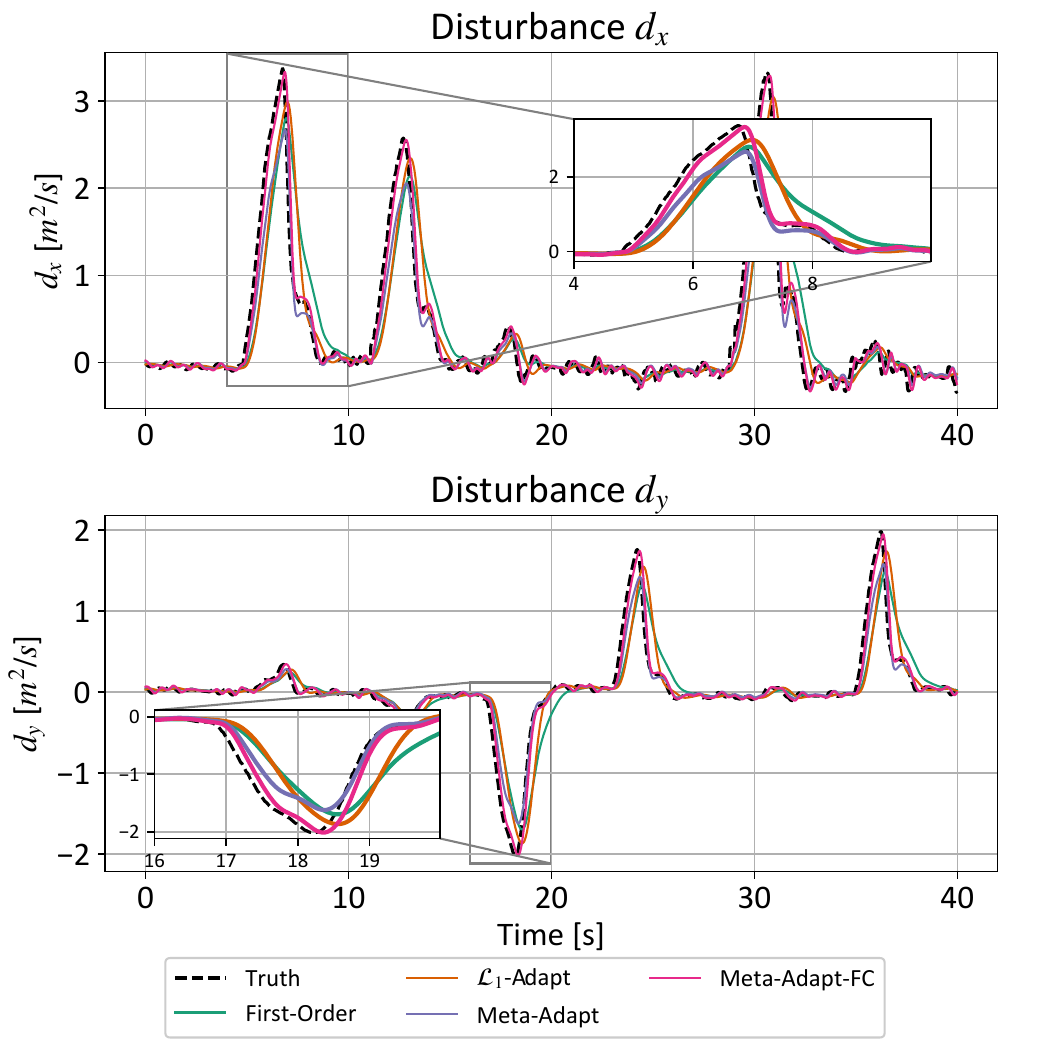}
  \vspace{-1.0em}
  \caption{The results of real-world push-rod force estimation in simulations.}
  \vspace{-1.0em}
  \label{fig:push_force_estimation}
\end{figure}

\begin{figure}[htbp]
  \centering
  \includegraphics[width=1.0\linewidth]{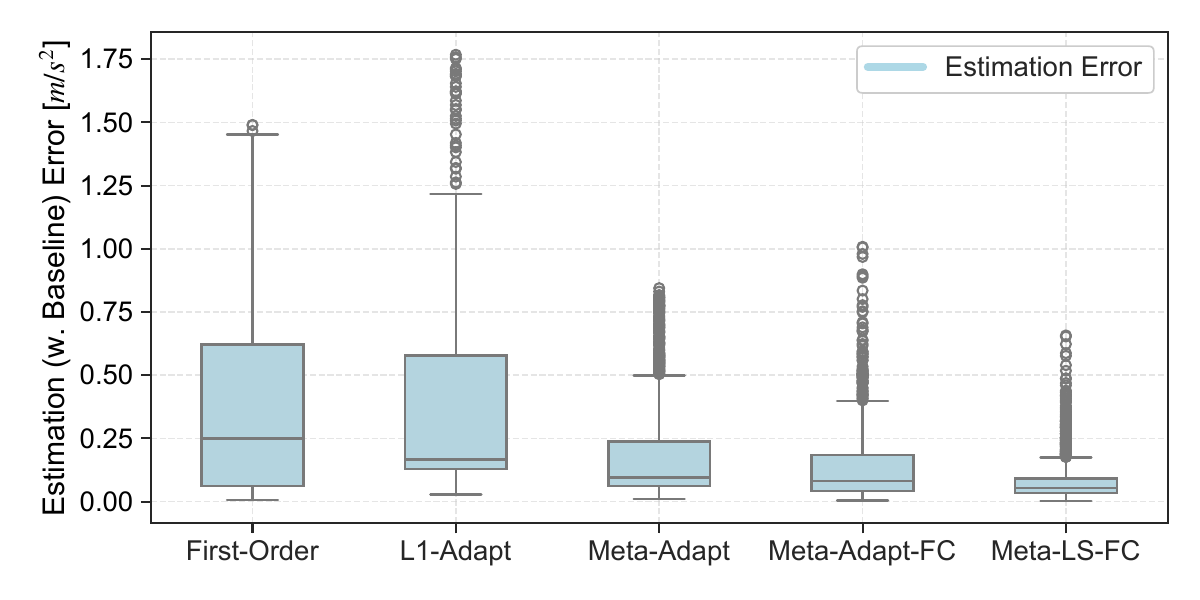}
  \vspace{-1.0em}
  \caption{The boxplots of real-world push-rod force estimation error.}
  \vspace{-1.0em}
  \label{fig:push_force_boxplot}
\end{figure}

\begin{figure*}[t!]
  \centering
  \includegraphics[width=1.0\textwidth]{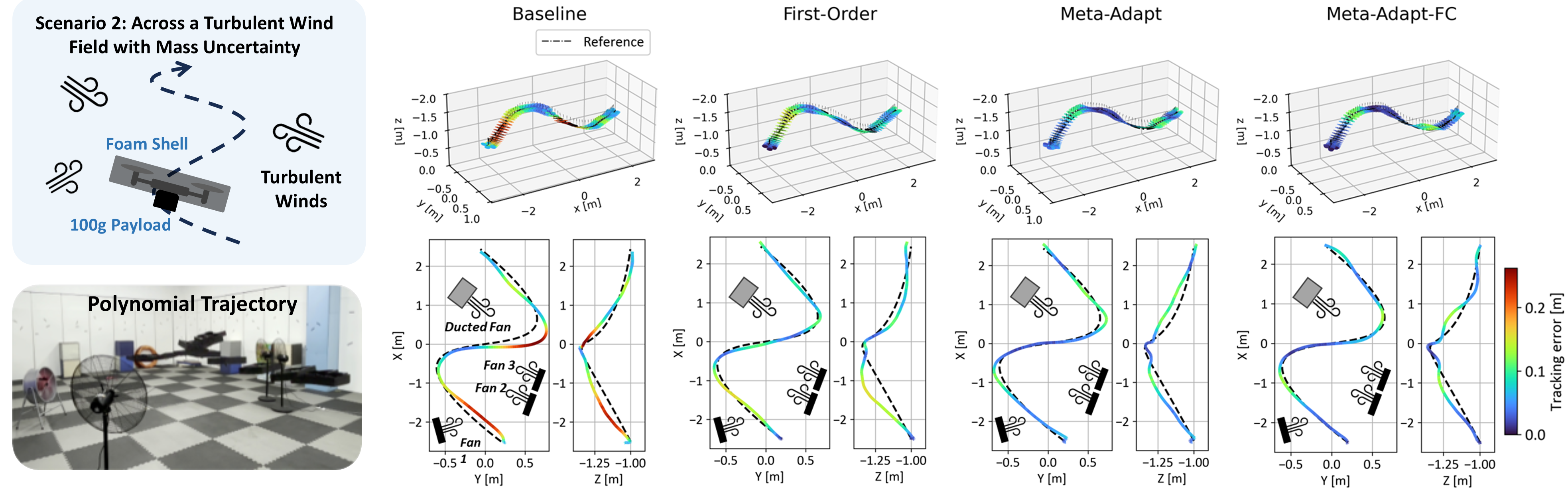}
  \vspace{-1.0em}
  \caption{Scenario.2, the quadrotor is commanded to fly across a turbulent wind field with mass uncertainty. 
  Trajectories are plotted with control error in 3D view and top view with the location of wind sources marked.}
  \vspace{-1.0em}
  \label{fig:exp2_general}
\end{figure*}

\subsection{Real-world Experiments}
In this section, the proposed framework is validated on a quadrotor platform
with three challenging tasks involving non-structural disturbances. Our approach
generalizes well to different scenarios using a single meta-representation, though the disturbance distribution
in the real world shifts away from the training dataset, as shown in Figure.\ref{fig:distribution}.
The quadrotor uses a motion capture system and onboard IMU for state and acceleration measurements, respectively.
For the details of the platform, controller, model parameters, and the disturbance estimation plots, see the
\href{https://nonstructural-metalearn.github.io/static/pdfs/apdx.pdf}{\textcolor{blue}{Supplementary Material IV}}.

\begin{figure}[t!]
  \centering
  \vspace{1.0em}
  \includegraphics[width=1.0\linewidth]{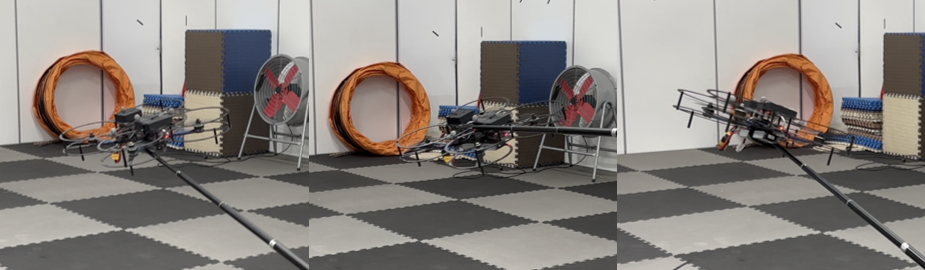}
  \vspace{-1.0em}
  \caption{The quadrotor is disturbed by a series of external forces via a push-rod.}
  \vspace{-1.0em}
  \label{fig:push_force_snapshot}
\end{figure}

\subsubsection{Estimation of Push-Rod Forces} 
The proposed method is first validated on a quadrotor platform
with a push-rod force estimation task.
The sensor measurements of sudden force injections are collected, followed by force estimation in simulations.
In this section, we provide the details of the external force estimation.
The quadrotor undergoes a series of external forces provided manually via a push-rod, as shown in Figure.\ref{fig:push_force_snapshot}.
The IMU measurement of acceleration is collected from the quadrotor,
which is then processed by an offline Gaussian Filter.
In the simulated environment, the quadrotor is commanded to hover at a fixed position
and the force is injected into the quadrotor in the earth-fixed frame.
The results of the estimation and the boxplots of the force estimation are provided in 
Figure.\ref{fig:push_force_estimation} and Figure.\ref{fig:push_force_boxplot}, respectively.
Our approach shows the least lagging in disturbance estimation, even when the disturbances
are completely state-independent and non-structural.

\subsubsection{Scenario.1: Fly with Suspended Payload and Aerodynamic Drag}
As shown in Figure.\ref{fig:exp1_general}, 
the quadrotor carries a suspended payload of $0.1\,\mathrm{kg}$ (about $12\%$ of its mass) 
and is wrapped with a foam shell to increase aerodynamic drag. 
We compare two meta-learned models with different regression lengths: 
one trained with $N=10$ and another with $N=20$ (denoted by the superscript \textbf{$^-$}). 
As shown in the boxplots in Figure~\ref{fig:exp1_general}, the unified meta-representation enables effective estimation of coupled disturbances, 
even though such conditions were not present during training. 
While the quality of the representation (the regression length) is crucial for the online adaptation, 
the feedback calibration mechanism further improves overall performance and robustness.
% Besides, an extra comparison on model performance has been conducted with fixed payloads, 
% we refer to Appendix.\ref{apdx:experiments_s1s2} for more details, together with the disturbance estimation plots.
The least-square method leads to the highest estimation performance but triggers an oscillation in tracking control.
This can be explained as the low-level attitude controller fails to respond to the rapid change of
desired acceleration due to the actuator delay.

\subsubsection{Scenario.2: Across a Turbulent Wind Field with Mass Uncertainty}
\begin{figure}[htbp]
  \centering
  \includegraphics[width=1.0\linewidth]{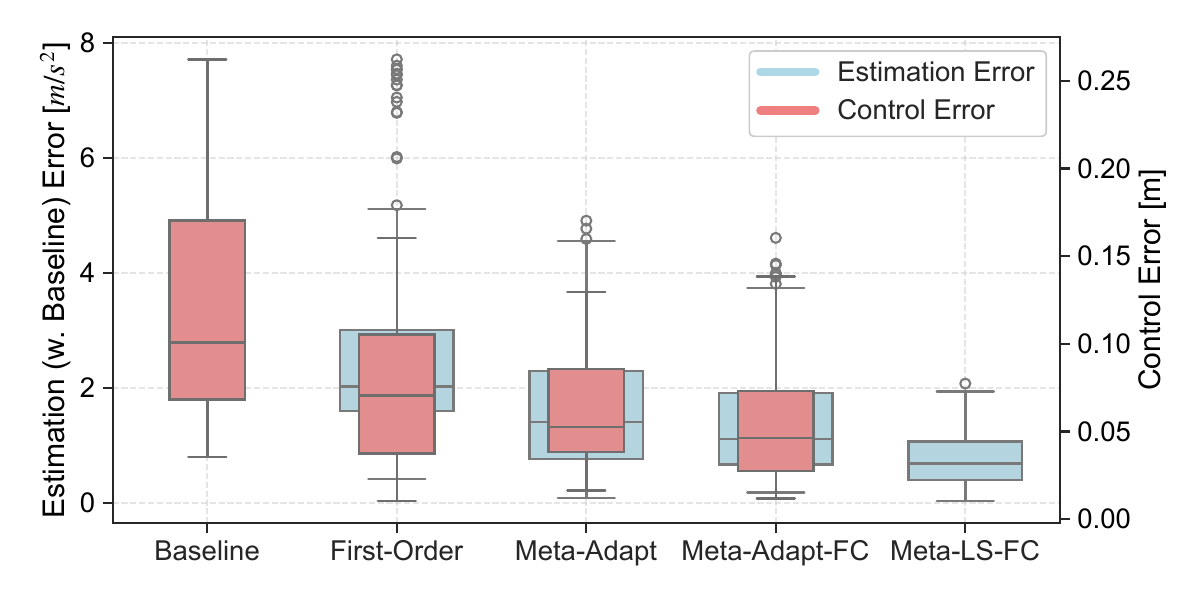}
  \vspace{-1.0em}
  \caption{Boxplots of estimation and control error in Scenario.2.}
  \label{fig:exp2_error_boxplot}
\end{figure}

This scenario tests the method with a quadrotor flying through a turbulent wind field while carrying a $0.1\,\mathrm{kg}$ payload and foam shell. 
As shown in Figure~\ref{fig:exp2_general}, the wind field is created by multiple fans with varying speeds and directions. 
The quadrotor tracks a polynomial trajectory across the wind field, and control errors are shown with color-coded lines.
As shown in Figure.\ref{fig:exp2_error_boxplot},
our method managed to estimate the complex gust disturbances with mass uncertainty, which was never seen during training.
The proposed feedback-calibration portion 
contributes to the model performance with a significant reduction in both the 
estimation error and control error.
The model adaptation performance is further improved with the least-square method in estimation,
the corresponding control task is not included due to potential oscillation in the strong winds.
Ablation study on the adaptation gain $\bm{P}$ can be found in 
\href{https://nonstructural-metalearn.github.io/static/pdfs/apdx.pdf}{\textcolor{blue}{Supplementary Material IV}}.

\section{Conclusion}
We proposed a feedback-calibrated meta-adaptive framework that enables
general non-structural disturbance prediction.
By combining the meta-learned model with a feedback-calibrated online adaptation mechanism, 
the proposed approach effectively captures general non-structural disturbances.
Theoretical analysis guarantees convergence of both model residuals and parameter estimation errors, 
while extensive simulations and real-world experiments on quadrotor platforms
validate the superiority of the method in trajectory tracking and disturbance rejection.
The proposed framework offers a generalizable solution for control and estimation 
of robotic systems in general disturbed environments,
bridging adaptive control and estimation from structured assumptions to more realistic generalizations.

\section{Limitations and Open Problems}
\textbf{The accessibility of acceleration.}
When acceleration measurements are unavailable, online adaptation can still proceed 
using the control error or state estimation error by augmenting 
the proposed framework into a state observer, where the latter one makes the whole framework
into a dual estimation one.

\textbf{Trade-off between regression length and online adaptation loss.}
The proposed method is capable of achieving a good trade-off between model performance and online adaptation loss.
The model performance is improved with a lower regression length $N$,
but the online adaptation loss is increased as the parameter estimator fails to track the optimal parameter in time. 
As in \cite{RSS2021_Adaptive-Control-Oriented}, an end-to-end learning approach that learns the representation 
under the dynamical effect of parameter estimation can be worth studying.

\textbf{The effect of the low-level controller.}
In real-world applications, the model performance is improved with the least-square method
but the tracking control performance is degraded.
The low-level controller can fail to respond to the rapid change of feedforward compensation.
Feedforward control that favors the low-level controller can be a potential future work.

\textbf{Extension to other robotic systems.}
Further extension can be made to more diverse robotic platforms and tasks, 
including legged locomotion and manipulation under dynamic contacts, 
where the environments are inherently more complex and frequent.

%===============================================================================

\bibliographystyle{IEEEtran}
\bibliography{reference}  % .bib
 
\end{document}